\documentclass[10pt,twocolumn,letterpaper]{article}

\usepackage{cvpr}
\usepackage{times}
\usepackage{epsfig}
\usepackage{graphicx}
\usepackage{amsmath}
\usepackage{amssymb}
\usepackage{float}

\iftrue     
\newcommand{\dimpp}[1]{\textcolor{blue}{[DP: #1]}}
\newcommand{\vic}[1]{\textcolor{blue}{[DP: #1]}}
\newcommand{\antonio}[1]{\textcolor{magenta}{[AT: #1]}}
\newcommand{\ferda}[1]{\textcolor{red}{[FO: #1]}}
\newcommand{\ingmar}[1]{\textcolor{cyan}{[IW: #1]}}
\newcommand{\youssef}[1]{\textcolor{green}{[YT: #1]}}
\else
\newcommand{\dimpp}[1]{\textcolor{blue}{\noindent}}
\newcommand{\vic}[1]{\textcolor{blue}{\noindent}}
\newcommand{\antonio}[1]{\textcolor{magenta}{\noindent}}
\newcommand{\ferda}[1]{\textcolor{red}{\noindent}}
\newcommand{\ingmar}[1]{\textcolor{cyan}{\noindent}}
\newcommand{\youssef}[1]{\textcolor{green}{\noindent}}
\fi

\newcommand{\mypar}[1]{\vspace{-4.2mm}\paragraph{#1}}

\newcommand{\norm}[1]{\left\lVert#1\right\rVert}

\usepackage[pagebackref=true,breaklinks=true,letterpaper=true,colorlinks,bookmarks=false]{hyperref}

\newcommand{\beforebeforeEquation}{\vspace{-2mm}}
\newcommand{\beforeEquation}{\vspace{-1.0mm}}
\newcommand{\afterEquation}{\vspace{-0.5mm}}

\cvprfinalcopy 

\def\cvprPaperID{3811} 

\ifcvprfinal\fi
\begin{document}

\title{How to make a pizza: \\ Learning a compositional layer-based GAN model}

\author{Dim~P.~Papadopoulos$^1$ \quad
Youssef Tamaazousti$^1$ \quad
Ferda Ofli$^2$ \quad
Ingmar Weber$^2$
\quad
Antonio Torralba$^1$ \\
$^1$ Massachusetts Institute of Technology \quad
$^2$ Qatar Computing Research Institute, HBKU
\\
{\tt\small\{dimpapa,ytamaaz,torralba\}@mit.edu, \{fofli,iweber\}@hbku.edu.qa}
}


\maketitle

\vspace{-2cm}
\begin{abstract}
   A food recipe is an ordered set of instructions for preparing a particular dish. From a visual perspective, every instruction step can be seen as a way to change the visual appearance of the dish by adding extra objects (e.g., adding an ingredient) or changing the appearance of the existing ones (e.g., cooking the dish). 
   In this paper, we aim to teach a machine how to make a pizza by building a generative model that mirrors this step-by-step procedure. To do so, we learn composable module operations which are able to either add or remove a particular ingredient.
   Each operator is designed as a Generative Adversarial Network (GAN). Given only weak image-level supervision, the operators are trained to generate a visual layer that needs to be added to or removed from the existing image.
   The proposed model is able to decompose an image into an ordered sequence of layers by applying sequentially in the right order the corresponding removing modules.
   Experimental results on synthetic and real pizza images demonstrate that our proposed model is able to: (1) segment pizza toppings in a weakly-supervised fashion, (2) remove them by revealing what is occluded underneath them (i.e., inpainting), and (3) infer the ordering of the toppings without any depth ordering supervision.
   Code, data, and models are available online\footnote{\url{http://pizzagan.csail.mit.edu}}.
   
\end{abstract}
\section{Introduction}

Food is an integral part of life that has profound implications for aspects ranging from health to culture. In order to teach a machine to ``understand'' food and its preparation, a natural approach is to teach it the conversion of raw ingredients to a complete dish, following the step-by-step instructions of a recipe. 
Though progress has been made on the understanding of the recipe-to-image mapping using multi-modal embeddings \cite{CarvalhoM:SIGIR18,salvador17cvpr}, remaining challenges include (i) the reconstruction of the correct steps in the recipe, and (ii) dealing with partial occlusion of ingredients for food that consists of different layers. 

\vspace{-0.5mm}
\begin{figure}[t]
\center
\includegraphics[width=1\linewidth]{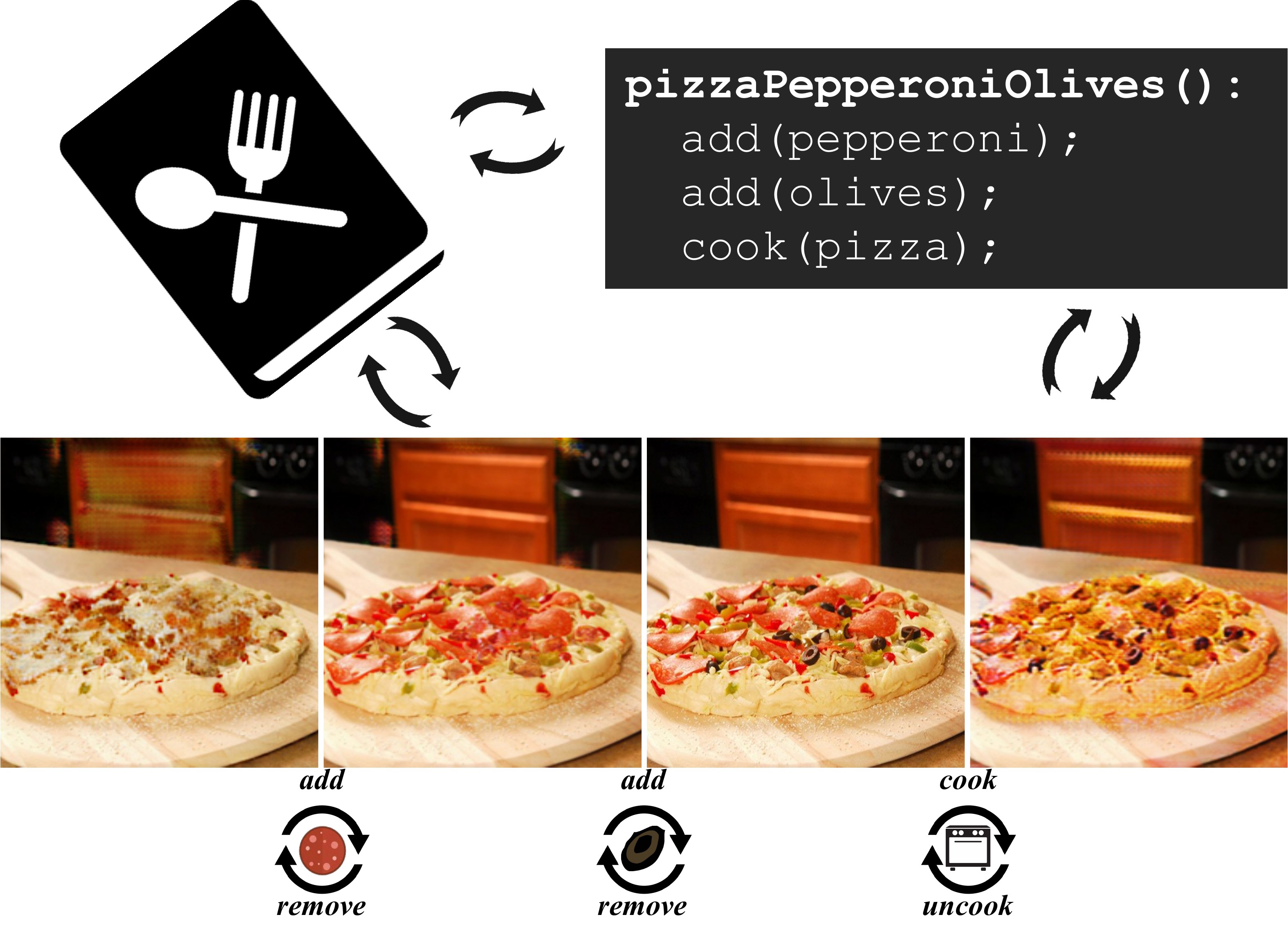} 
\vspace{-4.5mm}
\caption{\small \textbf{How to make a pizza:} We propose \emph{PizzaGAN}, a compositional layer-based generative model that aims to mirror the step-by-step procedure of making a pizza.
}
\vspace{-5mm}
\label{fig:splash}
\end{figure}

The archetypal example for this is pizza (Fig.~\ref{fig:splash} (left)).
%
The recipe for making a pizza typically requires sequentially adding several ingredients in a specific order on top of a pizza dough. This ordering of the adding operations defines the overlap relationships between the ingredients. In other words, creating this pizza image requires sequentially rendering different ingredient layers on top of a pizza dough image. Following the reverse procedure of sequentially removing  the ingredients in the reverse order corresponds to decomposing a given image into its layer representation (Fig.~\ref{fig:splash} (right)).
Removing an ingredient requires not only detecting all the ingredient instances but also resolving any occlusions with the ingredients underneath by generating the appearance of their invisible parts.
Going beyond food, the concept of ``layers''  is widespread in digital image editing, where images are composed by combining different layers with different alpha masks.
%



In this paper, we propose \emph{PizzaGAN}, a compositional layer-based generative model that mirrors this step-by-step procedure of making a pizza. 
Given a set of training images with only image-level labels (e.g., ``pepperoni pizza''), for each object class (e.g., ``pepperoni''), we learn a pair of module operators that are able to add and remove all instances of the target object class (e.g., ``add pepperoni'' and ``remove pepperoni''). Each such module operator is designed as a generative adversarial network (GAN). Instead of generating a complete new image, each adding GAN module is trained to generate (i) the appearance of the added layer and (ii) a mask that indicates the pixels of the new layer that are visible in the image after adding the layer. Similarly, each removing module is trained to generate (i) the appearance of the occluded area underneath the removed layer and (ii) a mask that indicates the pixels of the removed layer that will not be visible in the image after removing this layer. 

Given a test image, the proposed model can detect the object classes appearing in the image (\textit{classification}). Applying the corresponding removing modules sequentially results in decomposing the image into its layers.
%
We perform extensive experiments on both synthetic and real pizzas to demonstrate that our model is able to (1) detect and segment the pizza toppings in a weakly-supervised fashion without any pixel-wise supervision, (2) fill in what has been occluded with what is underneath (i.e., inpainting), and (3) infer the ordering of the toppings without any depth ordering supervision.

\begin{figure*}[t]
\center
\vspace{-2mm}
\includegraphics[width=\linewidth]{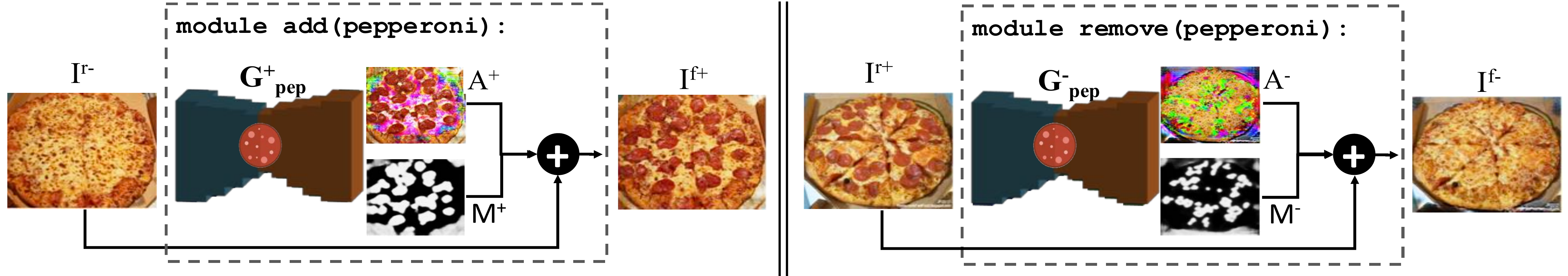}
\vspace{-3mm}
\caption{\small \textbf{Module operators that are trained to add and remove pepperoni on a given image.} Each operator is a GAN that generates the appearance $A$ and the mask $M$ of the adding or the removing layer. The generated composite image is synthesized by combining the input image with the generated residual image.}
\vspace{-5mm}
\label{fig:mainScheme}
\end{figure*}

\section{Related work}

\paragraph{Generative Adversarial Networks (GANs).} 
Generative Adversarial Networks (GANs)~\cite{arjovsky17arxiv,denton15nips,goodfellow14nips,radford16iclr,salimans16nips} are generative models that typically try to map an input random noise vector to an output image. GANs consist of two networks, a generator and a discriminator which are trained simultaneously. The generator is trained to generate realistic fake samples while the discriminator is trained to distinguish between real and fake samples. GANs have been used in various important computer vision tasks showing impressive results in image generation~\cite{huang17cvpr,radford16iclr}, image translation~\cite{choi18cvpr,isola17cvpr,liu17nips,zhu17iccv}, high quality face generation~\cite{karras18iclr}, super-resolution~\cite{ledig17cvpr}, video generation~\cite{denton18icml,tulyakov18cvpr,vondrick16nips}, video translation~\cite{bansal18eccv}, among many others.

\mypar{Image-to-image translation.} Conditional GANs (cGANs) are able to generate an output image conditioned on an input image. This makes these models suitable for solving image-to-image translation tasks where an image from one specific domain is translated into another domain. 
Several image-to-image translation approaches based on cGANs have been proposed~\cite{bousmalis17cvpr,choi18cvpr,huang18arxiv,isola17cvpr,liu17nips,perarnau16arxiv,wang18cvpr,zhu17iccv,zhu17nips}.
Isola et al.~\cite{isola17cvpr} proposed a generic image-to-image translation approach using cGANs trained with a set of aligned training images from the two domains. CycleGAN~\cite{zhu17iccv} bypasses the need of aligned pairs of training samples by introducing a cycle consistency loss that prevents the two generators from contradicting each other and alleviates the mode collapse problem of GANs.

In this paper, we formulate every object manipulation operator (e.g., add/remove) as an unpaired image-to-image translation and build upon the seminal work of CycleGAN~\cite{zhu17iccv}.
Our work offers extra elements over the above image-to-image translation approaches by building composable modules that perform different object manipulation operations, generating a layered image representation or predicting the depth ordering of the objects in the image.

\mypar{Image layers.}
Decomposing an image into layers is a task that was already addressed in the 90s
~\cite{darrell91visualmotion,he98layered,szeliski00cvpr,torr01pami,wang94tip}.
More recently, Yang et al.~\cite{yang12pami} proposed a layered model for object detection and segmentation that estimates depth ordering and labeling of the image pixels. In~\cite{yang17iclr},
the authors use the concept of image layers and propose a layered GAN model that learns to generate background and foreground images separately and recursively and then compose them into a final composite image.

Several approaches have been also proposed for the amodal detection~\cite{kar15iccv} or segmentation~\cite{follmann18arxiv,li2016amodal,zhu2017semantic}, the task of detecting or segmenting the full extent of an object including any invisible and occluded parts of it.
The recent work of Ehsani et al.~\cite{ehsani18cvpr} tries not only to segment invisible object parts but also to reveal their appearance.

\mypar{Generating residual images.} Recently, researchers have explored the idea of using a cGAN model to generate only residual images, i.e., only the part of the image that needs to be changed when it is translated into another domain, for the task of face manipulation~\cite{pumarola18eccv,shen17cvpr,zhao18eccv}.
For example, these models are able to learn how to change the hair color, open/close the mouth, or change facial expressions by manipulating only the corresponding parts of the faces. Instead, in this paper, we exploit the generation of residual images to infer a layer representation for an image. 

\mypar{Modular GAN.} Our work is also related to studies investigating the modularity and the composability of GANs~\cite{graesser18arxiv,zhao18eccv}. 
Recently, Zhao et al.~\cite{zhao18eccv} proposed a modular multi-domain GAN architecture, which consists of several composable modular operations. However, they assume that all the operations are order-invariant which cannot be true for adding and removing overlapping objects in an image. Instead, our model takes into account the layer ordering and is able to infer it at test time without any supervision.

\mypar{Image inpainting.} Removing an object from a natural image requires predicting what lies behind it by painting the corresponding pixels. Image inpainting, the task of reconstructing missing or deteriorated regions of an image, has been widely explored in the past by the graphics community~\cite{bertalmio00siggraph,criminisi04tip,Hays07}. Recently, several approaches using GANs have been proposed~\cite{pathak16cvpr,yang17cvpr,yeh17cvpr} to solve the task. The main difference of our removing modules is that one single GAN model is responsible for both segmenting the desired objects and generating the pixels beneath them.

\section{Method}

We now describe our proposed \emph{PizzaGAN} model. In this paper, we are given a set of training RGB images $\mathbf{I} \in  \mathbb{R}^{H \times W \times 3}$ of height $H$ and width $W$ with only image-level labels. Let $\mathbf{C}=\{c_1,c_2,...,c_k\}$ be the set of all the $k$ different labels (i.e., pizza toppings) in our pizza dataset.
For each training image $I_j$, we are given a binary vector of length $k$ that encodes the image-level label (i.e., topping) information for this image.

Our goal is to learn for each object class $c$ two mapping functions to translate images without any instance of class $c$ to images with instances of class $c$ (i.e., \textit{adding} class $c$) and vice versa (i.e., \textit{removing} class $c$). To do that, for each class $c$, we split the training samples into two domains: one with the images which contain the class $c$ ($X_c^+$) and one with the images that do not contain it ($X_c^-$). 

\subsection{Architecture of modules}

\paragraph{Generator Module.} Let $\mathbf{G}_c^{+}$ be the generator module that \textit{adds} a layer of the class $c$ on an input image $I^{r-}$ (mapping  $\mathbf{G}_c^+: X_c^- \rightarrow X_c^+$). Also, let $\mathbf{G}_c^{-}$ be the corresponding generator module that \textit{removes} the layer of the class $c$ (mapping  $\mathbf{G}_c^-: X_c^+ \rightarrow X_c^-$). This pair of generator modules is shown in Fig.~\ref{fig:mainScheme} for the class pepperoni.
Below, for simplicity we often omit the class $c$ from the notations.

The output generated images  $I^{f+} = \mathbf{G}^+(I^{r-})$ and $I^{f-} = \mathbf{G}^-(I^{r+})$ are given by:

\beforebeforeEquation
\begin{equation}
\beforeEquation
    I^{f+} = M^+ \odot A^+ + (1-M^+) \odot I^{r-}
\end{equation}
\vspace{-1.6em}
\begin{equation}
    I^{f-} = M^- \odot A^- + (1-M^-) \odot I^{r+}
\afterEquation
\end{equation}

\vspace{-1mm}


\noindent where $M^+, M^- \in [0,1]^{H \times W}$ are the layer masks that indicate how each pixel of the adding or the removing layer, respectively, will affect the final composite generated image. $A^+ \in \mathbb{R}^{H \times W \times 3}$ is the RGB image that captures the appearance of the adding layer, while $A^- \in \mathbb{R}^{H \times W \times 3}$ is the RGB image that captures the appearance of the parts that were occluded by the removing layer. In Fig.~\ref{fig:mainScheme}, we observe that $A^+$ captures the appearance of the pepperoni while $A^-$ captures the appearance of the cheese that lies underneath the pepperoni. 
The $\odot$ denotes the element-wise product.
Note that all the non-zero values of $M^+$ and $M^-$ denote the pixels that change in the output composite image.

\mypar{Discriminator.}
Our model contains a single discriminator $\mathbf{D}$, which is responsible for evaluating the quality of the generated composite images.
This network is trained to (i) distinguish whether the input image is real or fake ($D_{adv}$) and (ii) perform a multi-label classification task for the input image for all the classes ($D_{cls}$). 
These two objectives of the discriminator are crucial to force the generator to generate realistic images and, more importantly, to add or remove specific object classes from images without modifying the other class labels of the image.
%
Discriminator networks with an extra auxiliary classification output have been successfully used in various GAN models~\cite{choi18cvpr,odena16arxiv,pumarola18eccv}.

\begin{figure*}[t]
\center
\vspace{-2mm}
\includegraphics[width=\linewidth]{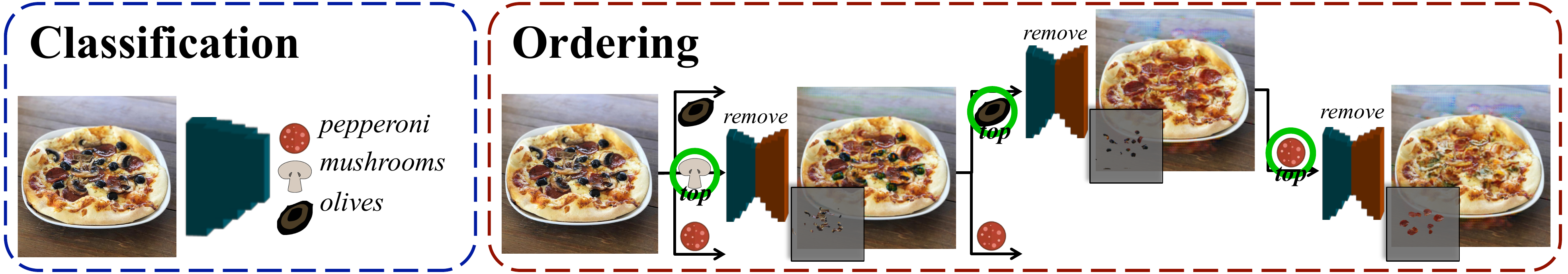}
\vspace{-2mm}
\caption{\small \textbf{Test time inference.} Given a test image, our proposed model detects first the toppings  appearing in the pizza (classification). Then, we predict the depth order of the toppings as they appear in the input image from top to bottom (ordering). The green circles in the image highlight the predicted top ingredient to remove.  Using this ordering, we apply the corresponding modules sequentially in order to 
reconstruct backwards the step-by-step procedure for making the input pizza.
}
\vspace{-5mm}
\label{fig:testTimeInfer}
\end{figure*}

\subsection{Learning the model}

All the adding $\mathbf{G}^+$ and removing $\mathbf{G}^-$ generator modules and the discriminator $\mathbf{D}$ are learned jointly. 
The full objective loss function contains four different terms: 
(a) \textbf{adversarial} losses that encourage the generated images to look realistic,
(b) \textbf{classification} losses that prevent the $\mathbf{G}^+$ and $\mathbf{G}^-$ to add or remove instances that belong to a different class than the target one,
(c) \textbf{cycle consistency} losses that prevent the $\mathbf{G}^+$ and $\mathbf{G}^-$ from contradicting each other,
(d) \textbf{mask regularization} losses that encourage the model to use both the generated layers and the input image.

\mypar{Adversarial loss.}

As in the original GAN~\cite{goodfellow14nips}, we use the adversarial loss to encourage the generated images to look realistic (i.e., match the distribution of the real image samples). For each adding module $\mathbf{G}^+$ and the discriminator $\mathbf{D}$, the adversarial loss is given by:

\vspace{-2mm}
\begin{equation}
    \begin{aligned}
        \mathcal{L}_{adv}(\mathbf{G}^+,\mathbf{D}) &= \mathbb{E}_{I^{r+}}[\log{D_{adv}(I^{r+})}] + \\
        & \quad \mathbb{E}_{I^{r-}}[\log{(1-D_{adv}(\mathbf{G}^+(I^{r-})))}]
    \end{aligned}
\end{equation}
\vspace{-2mm}

\noindent where $\mathbf{G}^+$ aims to generate realistic images, while $\mathbf{D}$ aims to distinguish between real images $I^{r+}$ and fake ones $I^{f+}$. $\mathbf{G}^+$ tries to minimize this loss, while $\mathbf{D}$ tries to maximize it. Similarly, we introduce an adversarial loss $\mathcal{L}_{adv}(\mathbf{G}^-,\mathbf{D})$ for each removing module $\mathbf{G}^-$ and the discriminator $\mathbf{D}$.

\mypar{Classification loss.}
As explained above, the discriminator $\mathbf{D}$ also performs a multi-label classification task. 
We introduce here a classification loss that encourages the generated images to be properly classified to the correct labels. This loss forces the generators to add or remove instances that belong only to the target class while preserving the class labels of all the other classes in the image. Without this loss, certain undesired effects occur such as the removal of class instances that should not be removed or the replacement of instances of an existing class when adding a new one.

This loss consists of two terms: a domain classification loss for the real images that we use to optimize $\mathbf{D}$, and a classification loss for the fake images that we use to optimize  $\mathbf{G}^+$ and $\mathbf{G}^-$. More formally we have:

\vspace{-1mm}
\begin{equation}
    \mathcal{L}_{cls}^r(\mathbf{D}) = \mathbb{E}_{I^{r}}[\norm{D_{cls}(I^{r})-l^r}^2] 
\end{equation}
\vspace{-4mm}
\begin{equation}
\begin{aligned}
    \mathcal{L}_{cls}^f(\mathbf{G}^+,\mathbf{G}^-,\mathbf{D}) &= \mathbb{E}_{I^{r-}}[\norm{D_{cls}(\mathbf{G}^+(I^{r-}))-l^{f+}}^2]+ \\
    & \quad  \mathbb{E}_{I^{r+}}[\norm{D_{cls}(\mathbf{G}^-(I^{r+}))-l^{f-}}^2]
\end{aligned}
\end{equation}
\vspace{-1mm}

\noindent where 
$D_{cls}$ represents a probability distribution
over all class labels computed by $\mathbf{D}$. $l^r$ represents a vector with the class level information of image $I^r$ while $l^{f+}$ and $l^{f-}$ represent the target class labels of the generated images.

\mypar{Cycle consistency loss.}

Using the adversarial and the classification losses above, the generators are trained to generate images that look realistic and are classified to the target set of labels. However, this alone does not guarantee that a generated image will preserve the content of the corresponding input image.
Similar to~\cite{zhu17iccv}, we apply a cycle consistency loss to the generators $\mathbf{G}^+$ and $\mathbf{G}^-$. The idea is that when we add something on an original image and then try to remove it, we should end up reconstructing the original image. More formally, we have:

\vspace{-3mm}
\begin{equation}
    \begin{aligned}
        \mathcal{L}_{cyc}^{I}(\mathbf{G}^+,\mathbf{G}^-) &= \mathbb{E}_{I^{r-}}[\norm{\mathbf{G}^-(\mathbf{G}^+(I^{r-})) - I^{r-}}_1] + \\
        &\quad  \mathbb{E}_{I^{r+}}[\norm{\mathbf{G}^+(\mathbf{G}^-(I^{r+}))- I^{r+}}_1] 
    \end{aligned}
\end{equation}
\vspace{-3mm}

The cycle consistency loss can be defined not only on the images but also on the generated layer masks. When we first add a layer and then remove it from an input image, the two generated layer masks $M^+$ and $M^-$ should affect in the same way the exact same pixels of the image. Similar to the above loss, we apply this second consistency loss in both directions:

\vspace{-3mm}
\begin{equation}
    \begin{aligned}
        \mathcal{L}_{cyc}^{M}(\mathbf{G}^+,\mathbf{G}^-) &= \mathbb{E}_{I^{r-}}[\norm{M^+(I^{r-}) - M^-(I^{f+})}_1] + \\
        &\quad  \mathbb{E}_{I^{r+}}[\norm{M^-(I^{r+}) - M^+(I^{f-})}_1] 
    \end{aligned}
\end{equation}
\vspace{-3mm}

Similar to~\cite{zhu17iccv}, we adopt the L1 norm for both cycle consistency losses. The final consistency loss ${L}_{cyc}$ for each pair of $\mathbf{G}^+$ and $\mathbf{G}^-$ is given by the sum of the two terms ${L}_{cyc}^{I}$ and ${L}_{cyc}^{M}$.

\mypar{Mask regularization.}
The proposed model is trained without any access to pixel-wise supervision, and therefore, we can not apply a loss directly on the generated masks $M$ and the appearance images $A$. 
These are learned implicitly by all the other losses that are applied on the final composite generated images. However, we often observe that the masks may converge to zero, meaning the generators have no effect. 
To prevent this, we apply a regularization loss on the masks $M^+$ and $M^-$:

\begin{equation}
    \begin{aligned}
    \mathcal{L}_{reg}(\mathbf{G}^+,\mathbf{G}^-) &= \mathbb{E}_{I^{r-}}[\norm{1-M^+(I^{r-})}_2] + \\
    &\quad \mathbb{E}_{I^{r+}}[\norm{1-M^-(I^{r+})}_2]
    \end{aligned}
\end{equation}

\mypar{Full loss.}
The full objective functions for the discriminator $\mathbf{D}$ and for each pair of adding and removing modules (i.e., $\mathbf{G}^+$ and $\mathbf{G}^-$) are defined as:


\begin{equation}
    \begin{aligned}
    \vspace{-3mm}
    \mathcal{L}_D &= -\sum_{c=1}^{k}{\mathcal{L}_{adv}(\mathbf{G}^+_c,\mathbf{D})} -
    \sum_{c=1}^{k}{\mathcal{L}_{adv}(\mathbf{G}^-_c,\mathbf{D})} + \\ 
    &\quad  \lambda_{cls}\sum_{c=1}^{k}{\mathcal{L}_{cls}^r(\mathbf{D})}
    \vspace{-2mm}
    \end{aligned}
\end{equation}
\vspace{-3mm}
\begin{equation}
    \begin{aligned}
    \mathcal{L}_{G_c} &= \mathcal{L}_{adv}(\mathbf{G}^+_c,\mathbf{D}) + 
    \mathcal{L}_{adv}(\mathbf{G}^-_c,\mathbf{D}) +  \\
    &\quad  \lambda_{cls}(\mathcal{L}_{cls}^{f}(\mathbf{G}^+_c,\mathbf{D}) + 
    \mathcal{L}_{cls}^{f}(\mathbf{G}^-_c,\mathbf{D})) + \\
    &\quad  \lambda_{cyc}(\mathcal{L}_{cyc}(\mathbf{G}^+_c,\mathbf{G}^-_c)) + 
    \lambda_{reg}(\mathcal{L}_{reg}(\mathbf{G}^+_c,\mathbf{G}^-_c))
    \end{aligned}
\end{equation}

\noindent where $\lambda_{cls}$, $\lambda_{cyc}$ and $\lambda_{reg}$ are hyper-parameters that control the relative importance of the classification loss, the cycle consistency loss, and the mask regularization loss compared to the adversarial loss.


\begin{figure}[t]
\center
\includegraphics[width=\linewidth]{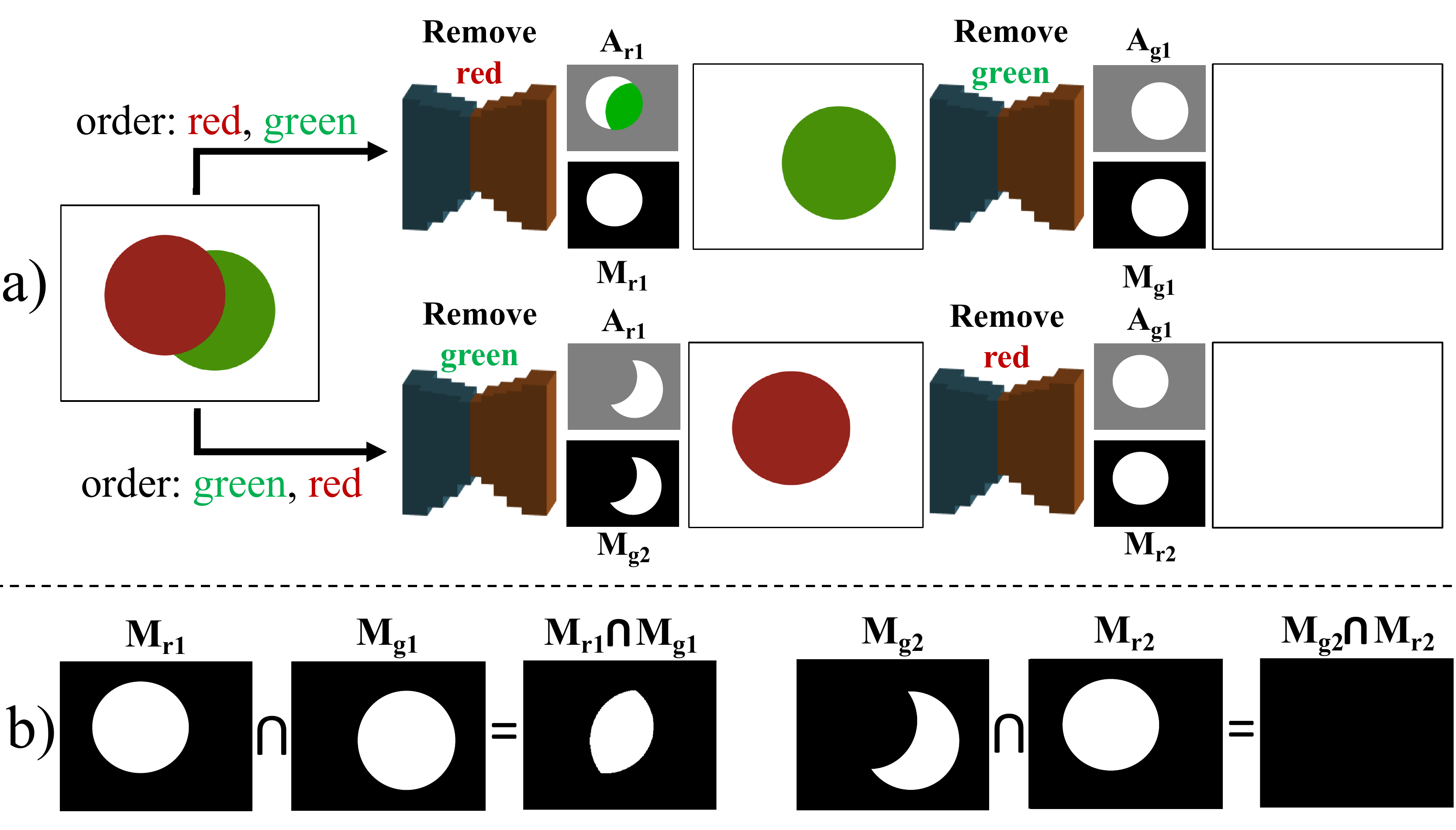}
\caption{\small \textbf{Predicting the ordering of layers}. (a) A toy example with two overlapping circular objects (red on top of green). In the first row we first remove the red object and then the green one, while in the second row we follow the reverse order. 
(b) Intersection between the two generated masks for each ordering permutation. 
We observe that in the first case the two generated masks $M$ highly overlap, while in the second one the overlap is zero.}
\label{fig:orderingToy}
\end{figure}

\subsection{Test time inference}
\label{sec:MethodInfer}

At test time, one can arbitrarily stack different composable adding modules and construct a specific sequence of operators. This leads to the generation of a particular sequence of layers which are rendered into an image to create new composite images. This can be seen as an abstraction of generating (making) a pizza image given an ordered set of instructions.

The reverse scenario is to predict the ordered set of instructions that was used to create an image. In other words, given a test image without any supervision, the goal here is to predict the sequence of removing operators that we can apply to the image to decompose it into an ordered sequence of layers. The inference procedure is shown in Fig.~\ref{fig:testTimeInfer} and is described below.

\mypar{Classification.}
We first feed the image into the discriminator to predict which toppings appear in the image, i.e. which removing modules should be applied.

\mypar{Ordering.}
An important question that arises here is which is the right order of applying these module operations to remove the layers. To answer this question we should infer which object is on top of which one. We exploit here the ability of the proposed model to reveal what lies underneath the removed objects. In particular, we use the overlaps of the generate dmasks to infer the ordering of the layers without any supervision.

\begin{figure}[t]
\center
\includegraphics[width=\linewidth]{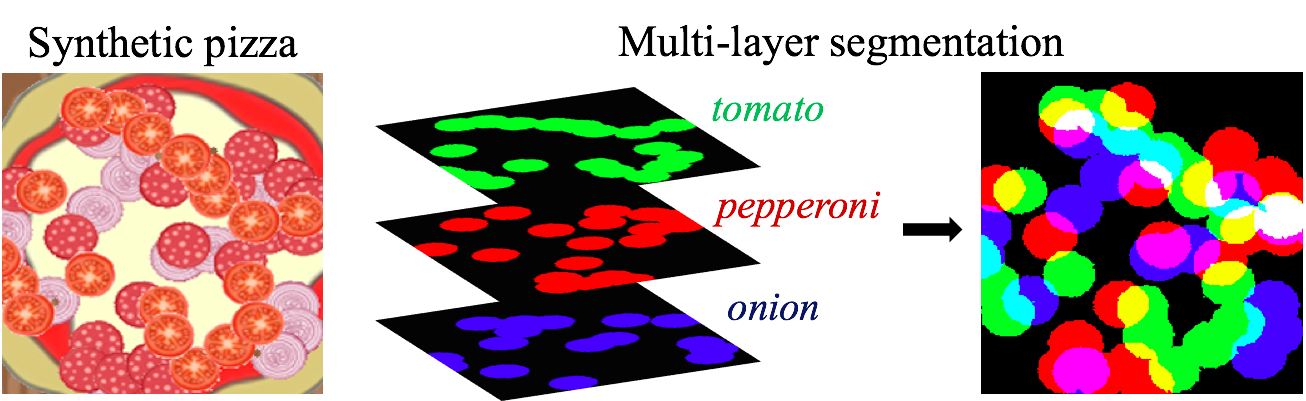}
\caption{\small \textbf{Ground-truth multi-layer segmentation of a synthetic pizza.} It captures all the occlusions that occur between the different toppings. For example, the cyan pixels (right) denote the parts of the onion occluded by a tomato.}
\vspace{-2mm}
\label{fig:syntheticGT}
\end{figure}


In Fig.~\ref{fig:orderingToy}, we use a toy example to explain the idea in more details. The image contains two overlapping circular objects with the red circle being on top of the green one. We investigate the two different permutations of ordering (red,green and green,red). 
We observe that in the first case the two generated masks of the modules highly overlap, while in the second case this overlap is zero (Fig.~\ref{fig:orderingToy}(b)). This happens because the model in the first case reveals green pixels below the red circle (see appearance image $A_{r1}$ in Fig.~\ref{fig:orderingToy}(a)) and completes the occluded green circle. Otherwise the resulting image will contain a green crescent (fake object) and not a green circle (real object). 
Therefore, we can predict the ordering between two objects by looking which ordering permutation leads to a higher overlap between the generated masks. 


In the general case of predicting the ordering between $m$ different layers, one should ideally try all different $m!$ ordering permutations. This is not feasible in practice.
Instead, we can still predict the full ordering by looking solely on the pairwise ordering between the $m$ layers. This results in only $m(m-1)$ pairwise permutations, making the ordering inference quite efficient. We also use the difference of the overlaps as an uncertainty measure to resolve contradictions in the pairwise predictions (e.g., $a$ on top of $b$, $b$ on top of $c$, $c$ on top of $a$) by simply ignoring the weakest pairwise ordering prediction (smallest difference of the overlaps).



\section{Collecting pizzas}

In this section, we describe how we create a synthetic dataset with clip-art-style pizza images (Sec.~\ref{sec:syntheticPizzas}) and how we collect and annotate real pizza images on Amazon Mechanical Turk (AMT) (Sec.~\ref{sec:realPizzas}).

\subsection{Creating synthetic pizzas}
\label{sec:syntheticPizzas}

There are two main advantages of creating a dataset with synthetic pizzas. First, it allows us to generate an arbitrarily large set of pizza examples with zero human annotation cost. Second and more importantly, we have access to accurate ground-truth ordering information and multi-layer pixel segmentation of the toppings. This allows us to accurately evaluate quantitatively our proposed model on the ordering and the semantic segmentation task. 
%
A ground-truth multi-layer segmentation for a synthetic pizza is shown in Fig.~\ref{fig:syntheticGT}. 
Note that in contrast to the standard semantic segmentation, every pixel of the image can take more than one object label (e.g., yellow pixels shown in Fig.~\ref{fig:syntheticGT} (right) denote the presence of both tomato and pepperoni).

\begin{figure}[t]
\center
\includegraphics[width=0.9\linewidth]{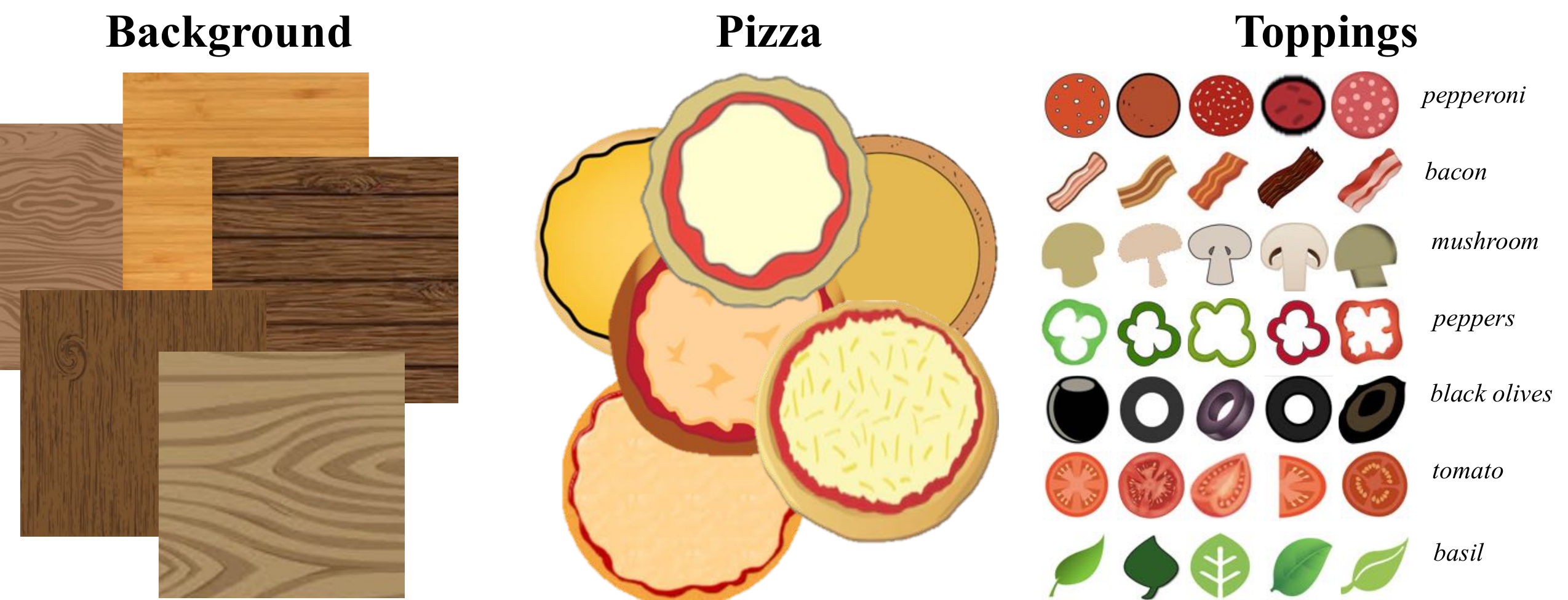} \\
\includegraphics[width=1\linewidth]{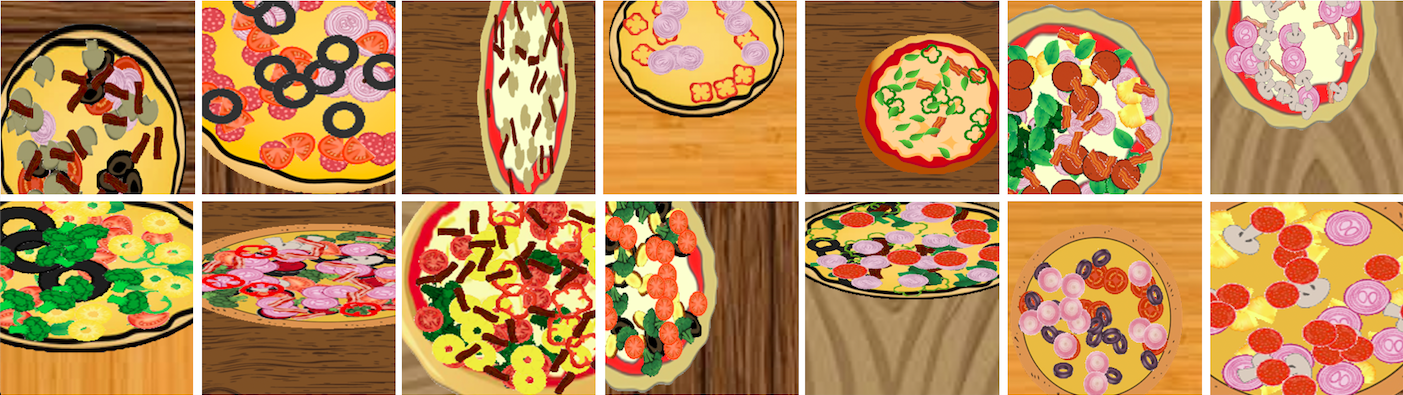}
\vspace{-3mm}
\caption{\small 
\textbf{Creating synthetic pizzas.} 
Top: Examples of background textures, base pizza images, and toppings used to create synthetic pizzas. Bottom: Examples of created synthetic pizzas.}
\vspace{-5mm}
\label{fig:baseImagesSynthetic}
\end{figure}

We use a variety of different background textures, different clip-art images of plain pizzas, and different clip-art images for each topping to obtain the synthetic pizzas (examples in Fig.~\ref{fig:baseImagesSynthetic} (top)). This adds a more realistic tone to the synthetic dataset and makes the task of adding and removing toppings a bit more challenging.
Examples of the obtained synthetic pizzas are shown in Fig.~\ref{fig:baseImagesSynthetic} (bottom). The dataset consists of pizzas with a wide variety of different configuration of the toppings (i.e., number of toppings, topping quantities, position of each instance of a topping, and ordering of topping layers).

\subsection{Collecting real pizzas}
\label{sec:realPizzas}


\paragraph{Data.}
Pizza is the most photographed food on Instagram with over 38 million posts using the hashtag $\#pizza$.
First, we download half a million images from Instagram using several popular pizza-related hashtags.
Then, we filter out the undesired images using a CNN-based classifier trained on a small set of manually labeled pizza/non-pizza images. 
%

\mypar{Image-level annotations.}
We crowd-source image-level labels for the pizza toppings on Amazon Mechanical Turk (AMT).
Given a pizza image, the annotators are instructed to label all the toppings that are visible on top of the pizza. 
%
Each potential annotator is first asked to complete a qualification test by annotating five simple pizza images. Qualification test is a common approach when crowd-sourcing image annotations as it enhances the quality of the crowd-sourced data by filtering out bad annotators~\cite{endres10eccv,Johnson11cvpr,krause2013iccvw,russakovsky15ijcv,sorokin08cvprw}.

Annotating pizza toppings can be challenging as several ingredients have similar visual appearances (e.g., bacon-ham, basil-spinach). To further ensure high quality, every image is annotated by five different annotators, and the final image labels are obtained using majority vote.

\begin{figure}[t]
\center
\includegraphics[width=\linewidth]{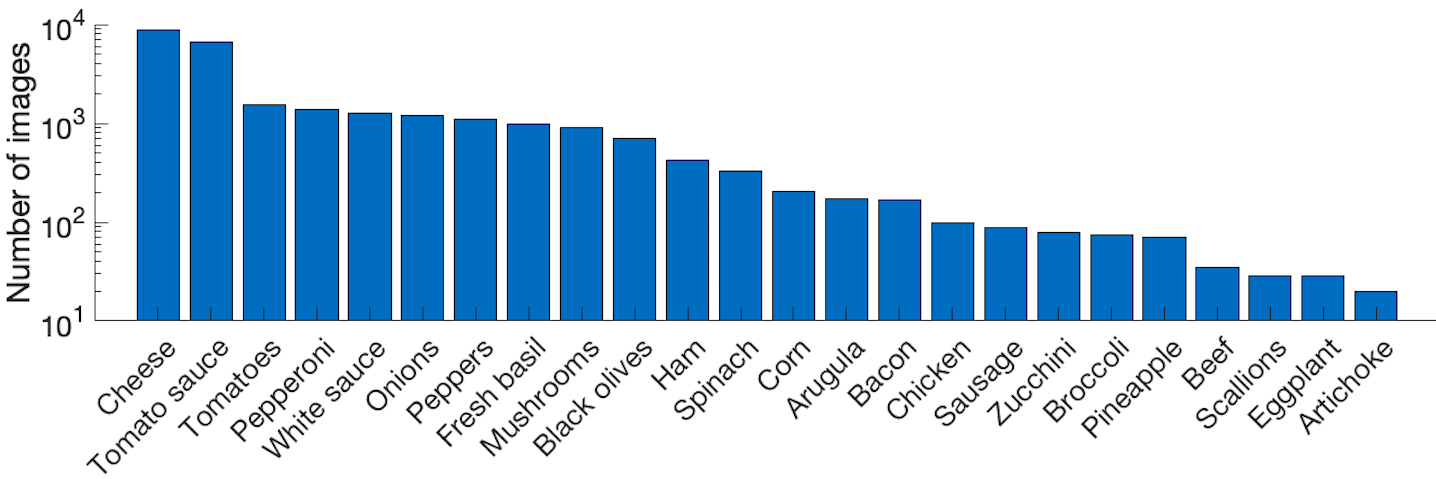}
\caption{\small The distribution of the toppings on the real pizzas.}
\vspace{-3mm}
\label{fig:distToppings}
\end{figure}

\mypar{Data statistics.}
Our dataset contains 9,213 annotated pizza images and the distribution of the labeled toppings is shown in Fig.~\ref{fig:distToppings}. The average number of toppings including cheese per pizza is 2.9 with a standard deviation of 1.1. 

\section{Implementation Details}

\paragraph{Architecture.}
The architectures of the generator modules and the discriminator are based on the ones proposed in~\cite{zhu17iccv}, as CycleGAN achieves impressive results on the unpaired image-to-image translation.
The generator architecture is modified by introducing an extra convolutional layer on top of the second last layer and in parallel with the existing last one. This layer has only one output channel and we use a sigmoid activation function for the mask.
For the discriminator, we adopt the popular PatchGAN architecture~\cite{isola17cvpr,zhu17iccv} that we slightly modify to also perform a multi-label classification task.

\mypar{Training details.}
We train our model using the Adam solver~\cite{kingma15iclr} with a learning rate of 0.0002 for the first 100 epochs. Then, we linearly decay it to zero over the next 100 epochs.
All generator modules and the discriminator are trained from scratch.
%
%
For all the experiments below, we set $\lambda_{cls}=1$, $\lambda_{cyc}=10$ and $\lambda_{reg}=0.01$.
For real pizzas, we first get a centered squared crop of the input images and then resize them to 256$\times$256.

\section{Experimental results}

\begin{figure}[t]
\center
\vspace{-0.5mm}
\includegraphics[width=\linewidth]{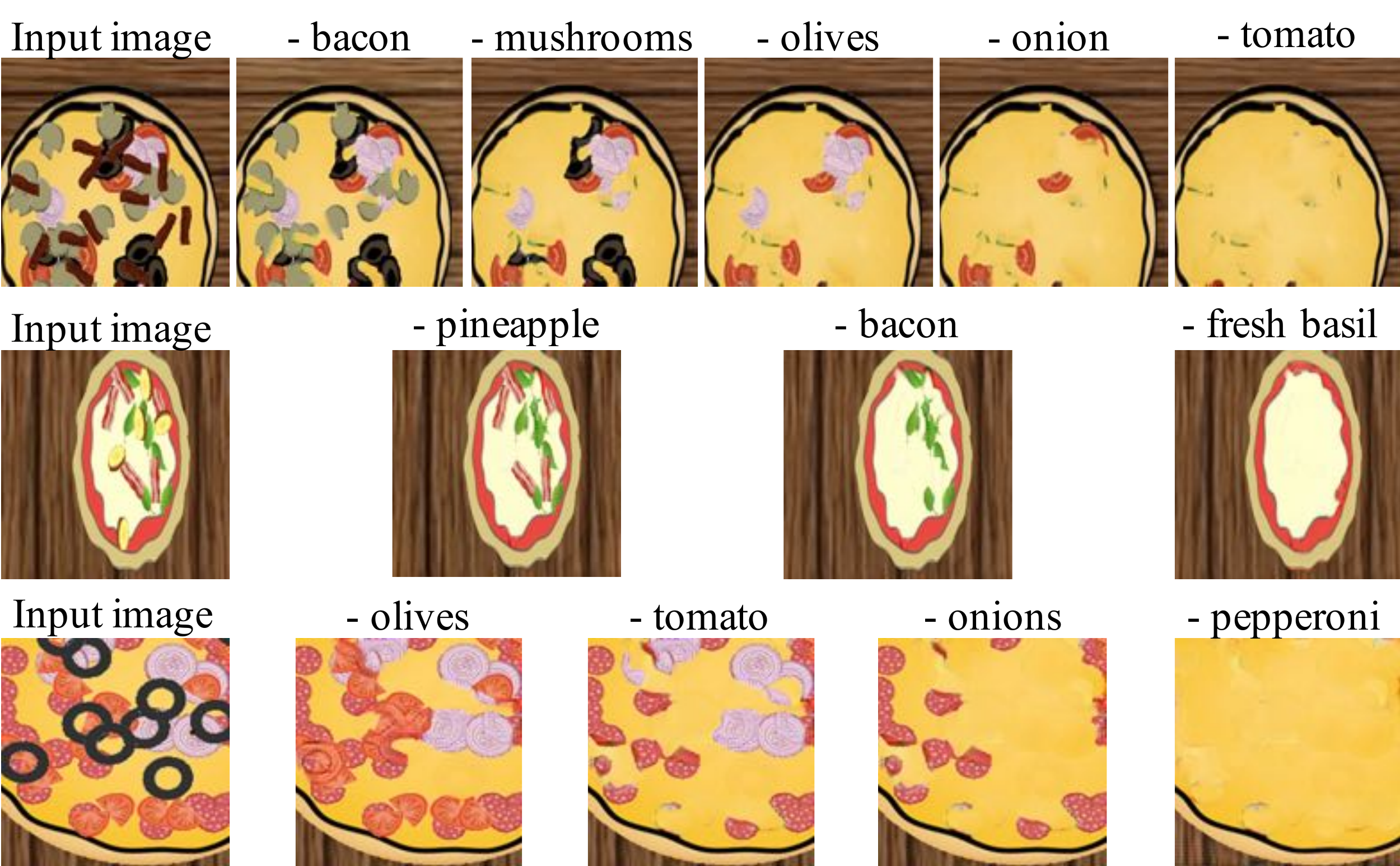}
\vspace{-3mm}
\caption{\small We predict the sequence of removing operators and apply them sequentially to the input image. Note how every time the current top ingredient is the one removed. This process reveals several invisible parts of ingredients when removing the top layers that occlude them.}
\vspace{-5mm}
\label{fig:qualFake}
\end{figure}

\subsection{Results on synthetic pizzas}

\paragraph{Data.}
We create a dataset of 5,500 synthetic pizzas. 
Each pizza can contain up to 10 toppings from the following list: 
\textit{\{bacon, basil, broccoli, mushrooms, olives, onions, pepperoni, peppers, pineapple, tomatoes\}}.
We split the images into 5,000 training and 500 test images.
We use the training images with accompanying image-level labels to train our model, and measure its performance on the test set.

\mypar{Qualitative results.} Fig.~\ref{fig:qualFake} shows qualitative results on  synthetic test images. We show how we can predict how a pizza was made: we predict the sequence of removing operators that we can apply to the image to decompose it into an ordered sequence of layers. 

\mypar{Evaluation.} 
Below, we evaluate our model on the following tasks: (i) multi-label topping classification, (ii) layer ordering prediction (see Sec.~\ref{sec:MethodInfer}), and (iii) weakly-supervised semantic segmentation.

We measure classification performance using mean average precision (mAP).
We quantify ordering accuracy using the Damerau–Levenshtein distance (DL) as the minimum number of operations (insertion, deletion, substitution, or transposition) required to change the ground-truth ordering into the predicted ordering normalized by the number of ground-truth class labels. 
We compute segmentation accuracy using the standard Intersection-over-Union (IoU) measure and, then, calculate the mean over all classes (mIoU).

\mypar{Classification.}
The classification of the toppings on the synthetic pizzas is a simple task. Our model achieves 99.9\% mAP. As a reference, a CNN classifier based on ResNet18~\cite{he16cvpr} trained from scratch using a binary cross-entropy loss achieves 99.3\% mAP. 

\mypar{Ordering.} The average normalized DL distance for our PizzaGAN is $0.33$. As a reference, a random sequence of random labels achieves $0.91$ while a random permutation of the oracle labels achieves $0.42$. These numbers express normalized distances, so a lower value indicates higher accuracy. We also evaluate the ordering accuracy only on a subset of the test images that contain exactly two toppings. We find that our method is able to predict the correct ordering 88\% of the times.

\begin{table}[tb] 
\centering
\bgroup
\def\arraystretch{1.1}
\resizebox{0.9\linewidth}{!}{
\begin{tabular}{|l c c|}
\hline
\textbf{Method} & Architecture & mIoU (\%)\\ 
\hline
\hline
\textbf{CAM~\cite{zhou16learning}}         & Resnet18~\cite{he16cvpr}  & 22.8  \\
\textbf{CAM~\cite{zhou16learning}}         & Resnet38+~\cite{zifeng16arxiv}  & 39.9  \\
\textbf{AffinityNet~\cite{ahn_CVPR18}}         & Resnet38+~\cite{zifeng16arxiv}  & 48.2  \\
\textbf{CAM~\cite{zhou16learning}+CRF}     & Resnet38+~\cite{zifeng16arxiv}  & 51.5  \\
\textbf{AffinityNet~\cite{ahn_CVPR18}+CRF}           &  Resnet38+~\cite{zifeng16arxiv}   & 47.8    \\
\hline
\textbf{PizzaGAN (no ordering)}                   &   & \textbf{56.7} \\
\textbf{PizzaGAN (with ordering)}                   &   & \textbf{58.2}  \\
\hline
\end{tabular}
}
\vspace{+0.5mm}
\caption{
\textbf{Weakly-supervised segmentation mIoU performance on synthetic pizzas.}
All comparison methods are pre-trained on ILSVRC, while PizzaGAN is trained from scratch. 
In Resnet38+, GAP and FC layers replaced by three atrous convolutions~\cite{chen2018deeplab}.  
}
\vspace{-5mm}
\label{tab:WSsegRes_synthPizzas}
\egroup
\end{table}

\mypar{Segmentation.}
We compare the segmentation masks generated by the removing modules with various weakly-supervised segmentation approaches (Tab.~\ref{tab:WSsegRes_synthPizzas}).

Class Activation Maps (CAMs)~\cite{zhou16learning} achieve 22.8\% using a ResNet18 architecture. ResNet18 has roughly the same number of parameters with the architecture of our generators. CAMs with a deeper network achieve 39.9\%.
AffinityNet~\cite{ahn_CVPR18} is a powerful CNN that builds upon CAMs and achieves state-of-the-art results on the PASCAL VOC 2012 dataset. Even though AffinityNet improves CAMs by about 8\%, our method outperforms it by $\sim10\%$. 
When applying denseCRFs on top of the predicted segments of CAMs and AffinityNet, the performance reaches 51.5\% mIoU, which is notably below the performance of our model. 
\begin{figure}[t]
\center
\vspace{-1mm}
\includegraphics[width=\linewidth]{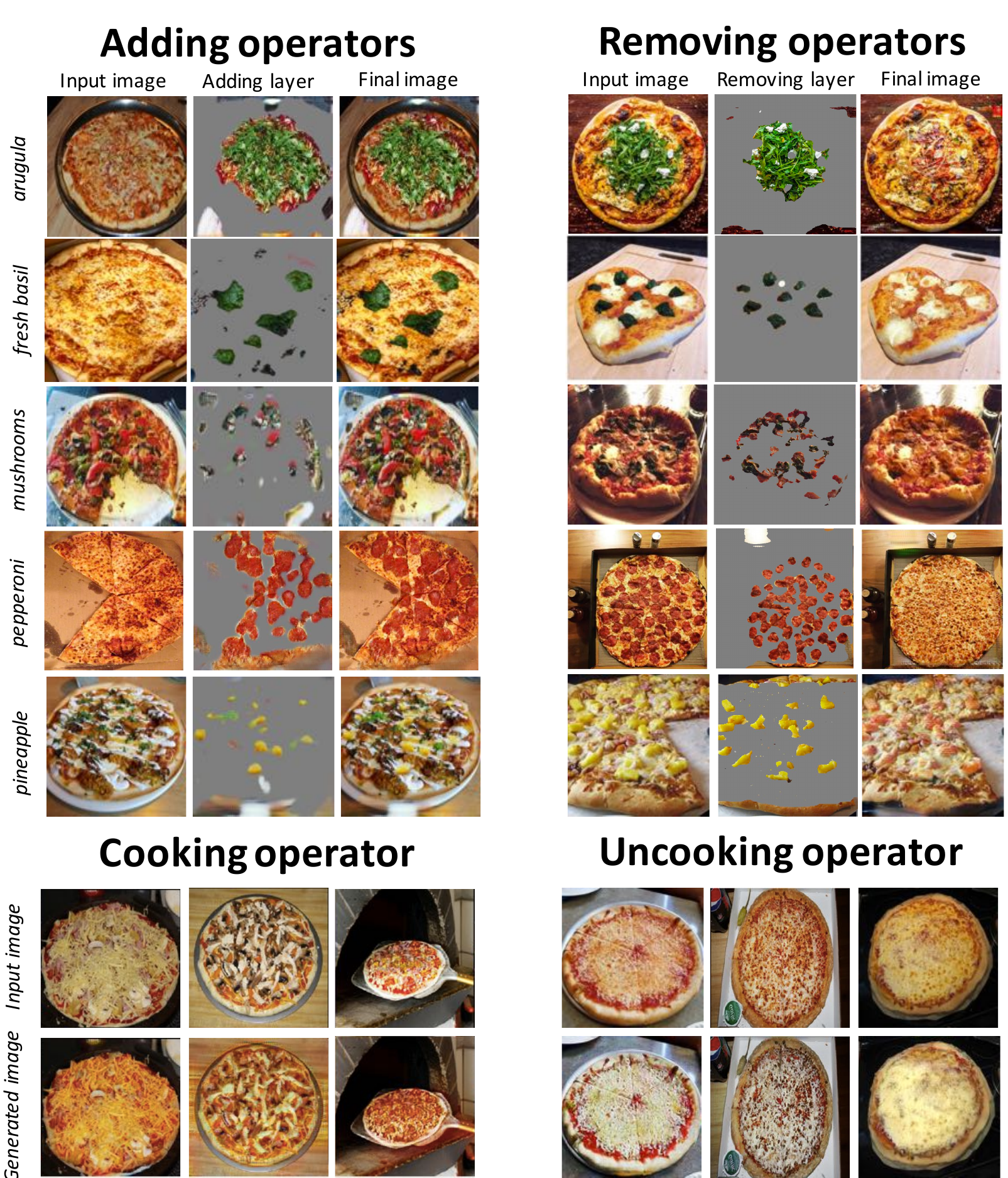} 
\caption{\small \textbf{Qualitative results of individual operators on real pizzas.} (top) Adding and removing operators. (bottom) Cooking and uncooking operators.}
\vspace{-3mm}
\label{fig:qualRealSolo}
\end{figure}

Our proposed PizzaGAN without any ordering inference (applying the removing modules in parallel on the input image) achieves 56.7\% mIoU. Applying the removing modules sequentially (based on the predicting ordering) brings an additional $+2\%$ in mIoU. This reflects the ability of our model to reveal invisible parts of the ingredients by removing the top ones first. Interestingly, using an oracle depth ordering, we achieve 60.9\% which is only 3\% higher than using the predicted ordering. This upper bound (oracle ordering) provides an alternative way to evaluate the impact of the depth ordering on the segmentation task.

\mypar{Occluded and non-occluded regions.}
To further investigate the impact of the ordering, we measure the segmentation performance broken into the occluded and non-occluded regions of the image. Without any ordering prediction, we achieve 70.4\% mIoU on the non-occluded regions and 0\% mIoU on the occluded ones. Using the predicted depth ordering, we achieve similar performance (70.5\%) on the non-occluded regions and 18.2\% on the occluded ones. This breakdown shows that the depth ordering enables the prediction of the occluded and invisible parts of the objects which can be very useful for various food recognition applications. 


\begin{figure}[t]
\center
\vspace{-2mm}
\includegraphics[width=\linewidth]{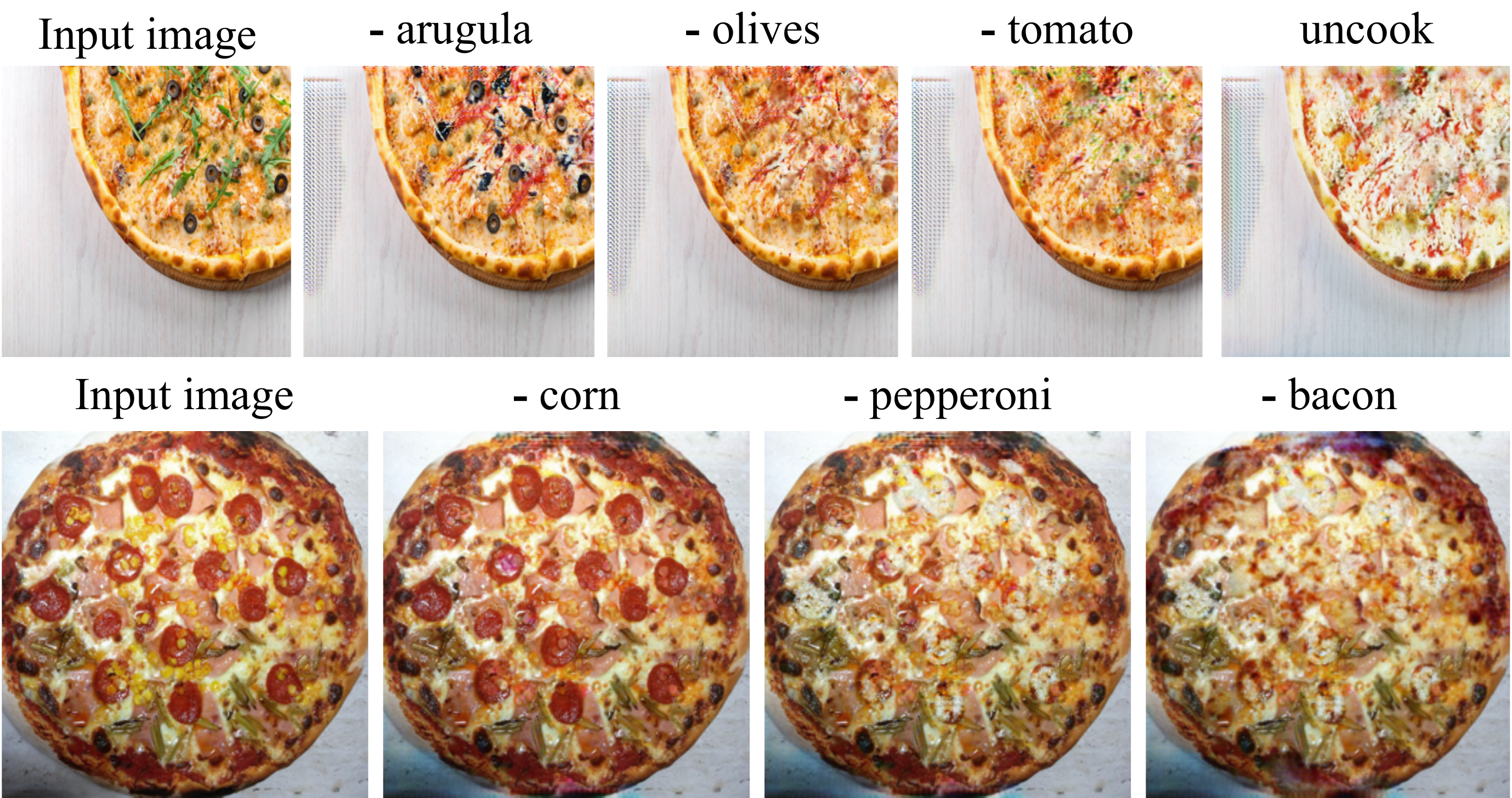}
\vspace{-3mm}
\caption{\small \textbf{PizzaGAN}: We predict the sequence of the operaors and apply them sequentially to the input image. The goal of the model is to remove every time the top ingredient. This leads in reconstructing backwards the recipe procedure used to make the input pizza.}
\vspace{-5mm}
\label{fig:sequence}
\end{figure}

\subsection{Experiments on real pizzas}

\begin{figure}[t]
\center
\vspace{-2mm}
\includegraphics[width=\linewidth]{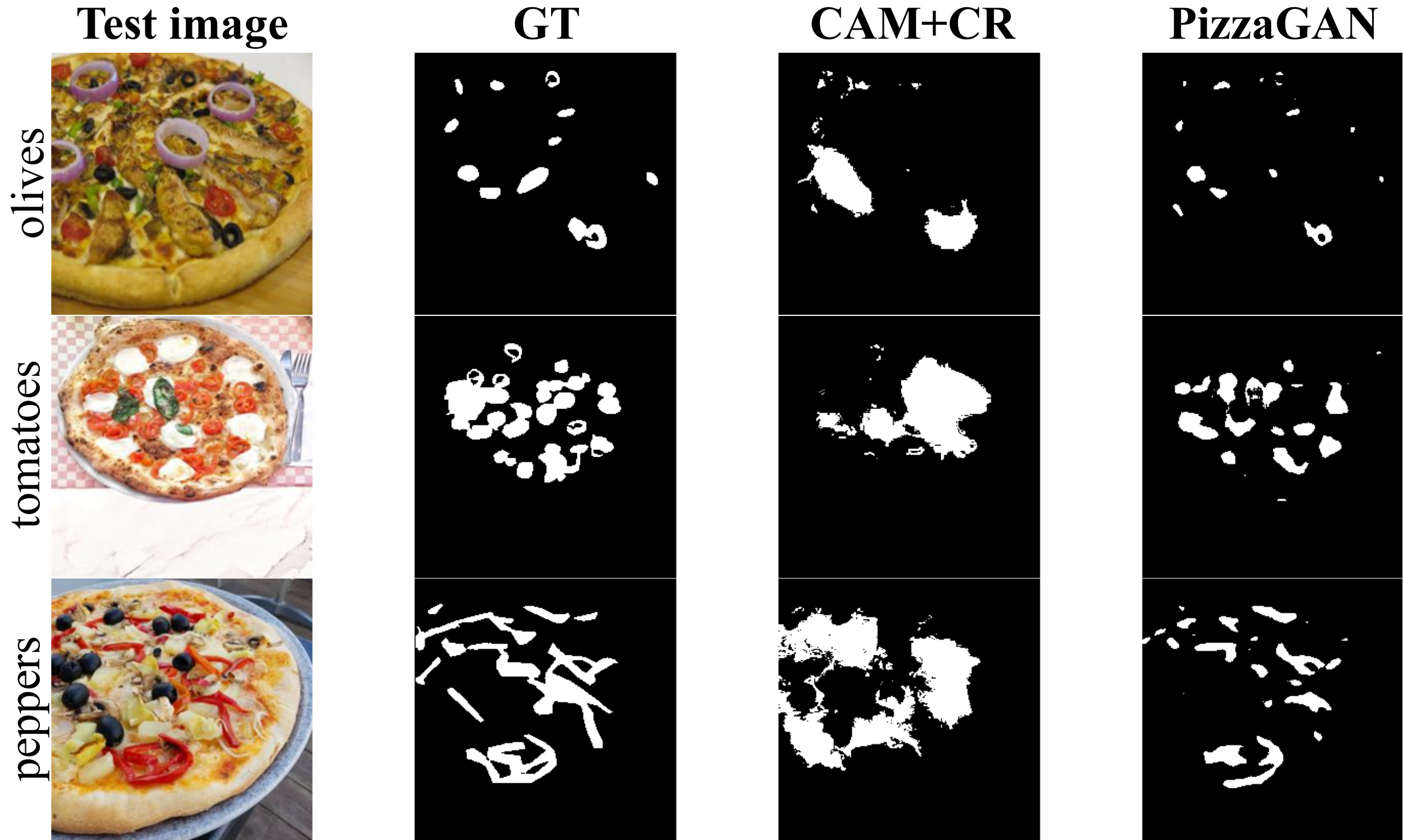}
\vspace{-3mm}
\caption{\small Examples of segmentation results on real pizzas. }
\vspace{-5mm}
\label{fig:segmReal}
\end{figure}

\paragraph{Data.}
In this section, we perform experiments on real pizzas.
We train our model on 9,213 images for 12 classes (toppings):
\textit{\{arugula, bacon, broccoli, corn, fresh basil, mushrooms, olives, onions, pepperoni, peppers, pineapple, tomatoes\}}.
For evaluation purposes, we manually annotate a small set of 50 images with accurate segmentation mask (see ground-truth segmentations in Fig.~\ref{fig:segmReal}). 
We use the same evaluation setup as for the synthetic pizzas and assess the classification and the weakly-supervised semantic segmentation performance. 

\mypar{Qualitative results.}
Fig.~\ref{fig:qualRealSolo}(top) shows the effect of individual adding and removing modules on real images. 
We observe that the adding modules learn \emph{where} to add by detecting the pizza and \emph{how} to add by placing the new pieces in a uniform and realistic way on the pizza. The removing modules learn \emph{what} to remove by accurately detecting the topping and  \emph{how} to remove by trying to predict what lies underneath the removed ingredient. 
In Fig.~\ref{fig:sequence} we show how we can predict how a pizza was made: we predict the sequence of operators that we can apply to the image to decompose it into an ordered sequence of layers.



\mypar{Classification.} Our model achieves 77.4\% mAP. As a reference, a CNN classifier based on ResNet18~\cite{he16cvpr} trained from scratch using a binary cross-entropy loss achieves 77.6\% mAP. 

\mypar{Segmentation.}
Our approach without any ordering inference (applying the removing modules in parallel on the input image) achieves 28.2\% mIoU. Applying the removing modules sequentially using the predicted ordering achieves 29.3\% mIoU. As expected the performance is significantly below the one observed on the synthetic data, given that real images are much more challenging than the synthetic ones.

Using ResNet38+, CAMs~\cite{zhou16learning} achieve 14.2\% and when applying a dense CRF on top of the predictions they achieve 22.7\%. 
Our proposed model outperforms both these models by a large margin ($+6.6\%$). 
Fig.~\ref{fig:segmReal} shows some segmentation prediction examples from CAMs+CRF and our PizzaGAN. 

\mypar{Cooking modules.} Besides adding and removing operations, the process of cooking a pizza is essential in the recipe procedure. We manually label here a subset of 932 pizzas as being cooked or uncooked in order to train modules that aim to cook or uncook a given pizza. The modules are trained similarly to the adding/removing ones and some qualitative results are shown in Fig.~\ref{fig:qualRealSolo}(bottom).

\section{Conclusions}
In this paper, we proposed \emph{PizzaGAN}, a generative model that mirrors the pizza making procedure. 
To this end, we learned composable module operations (implemented with GANs) to add/remove a particular ingredient or even cook/uncook the input pizza. 
In particular, we formulated the layer decomposition problem as several sequential unpaired image-to-image translations. 
%
Our experiments on both synthetic and real pizza images showed that our model (1) detects and segments the pizza toppings in a weakly-supervised fashion without any pixel-wise supervision, (2) fills in what has been occluded with what is underneath, and (3) infers the ordering of the toppings without any depth ordering supervision.

Though we have evaluated our model only in the context of pizza, we believe that a similar approach is promising for other types of foods that are naturally layered such as burgers, sandwiches, and salads.
Beyond food, it will be interesting to see how our model performs on domains such as digital fashion shopping assistants, where a key operation is the virtual combination of different layers of clothes.

{\small
\bibliographystyle{ieee}
\bibliography{shortstrings.bib,dimBibTex.bib}
}

\end{document}